\documentclass{article}

\usepackage{arxiv}

\usepackage[utf8]{inputenc} 
\usepackage[T1]{fontenc}    
\usepackage{hyperref}       
\usepackage{url}            
\usepackage{booktabs}       
\usepackage{amsfonts}       
\usepackage{nicefrac}       
\usepackage{microtype}      
\usepackage{lipsum}		
\usepackage{graphicx}
\usepackage{doi}
\usepackage{multirow}
\usepackage{colortbl}
\usepackage{dsfont}
\usepackage{amsmath,amssymb}
\usepackage{algorithm}
\usepackage{algorithmic}
\usepackage{tabularx}
\usepackage{stfloats}
\usepackage{array}
\usepackage{subcaption}
\usepackage{caption}

\title{Land Cover Image Classification}

\author{
\href{https://orcid.org/0009-0009-1399-7886}{\includegraphics[scale=0.06]{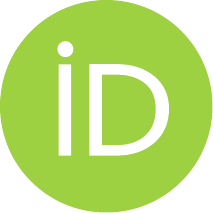}\hspace{1mm}Antonio Rangel} \\
    Instituto Tecnológico de Mazatlan\\	
    ITM\\
	\texttt{hectorarangel7@gmail.com} \\
\And
\href{https://orcid.org/0000-0001-6662-0390}{\includegraphics[scale=0.06]{orcid.pdf}\hspace{1mm}Juan R.~Terven} \\
    Instituto Politecnico Nacional\\	
    CICATA-Qro\\
	\texttt{jrtervens@ipn.mx} \\
\And
\href{https://orcid.org/0000-0002-5657-7752}{\includegraphics[scale=0.06]{orcid.pdf}\hspace{1mm}Diana M.~Cordova-Esparza} \\
Universidad Autónoma de Querétaro\\
Facultad de Informática\\
\texttt{diana.cordova@uaq.mx} \\
\And
\href{https://orcid.org/0000-0002-1938-8610}{\includegraphics[scale=0.06]{orcid.pdf}\hspace{1mm}E.A. Chávez-Urbiola} \\
    Instituto Politecnico Nacional\\	
    CICATA-Qro\\
	\texttt{eachavezu@ipn.mx} \\
}

\renewcommand{\headeright}{Published in the $1^{st}$ Conference in Technology and Applied Science 2023 }
\renewcommand{\undertitle}{Published in CITCA 2023}
\renewcommand{\shorttitle}{}

\hypersetup{
pdftitle={Land Cover Image Classification},
pdfsubject={cs, cs},
pdfauthor={Antonio Rangel, Juan R.~Terven, Diana M.~Cordova-Esparza, E.A. Chavez-Urbiola},
pdfkeywords={remote sensing, land cover, LULC classification},
}

\begin{document}
\maketitle

\begin{abstract}
Land Cover (LC) image classification has become increasingly significant in understanding environmental changes, urban planning, and disaster management. However, traditional LC methods are often labor-intensive and prone to human error. This paper explores state-of-the-art deep learning models for enhanced accuracy and efficiency in LC analysis. We compare convolutional neural networks (CNN) against transformer-based methods, showcasing their applications and advantages in LC studies. We used EuroSAT, a patch-based LC classification data set based on Sentinel-2 satellite images and achieved state-of-the-art results using current transformer models. 
\end{abstract}


\section{Introduction}
Land Use Land Cover (LULC) is a multidisciplinary field that categorizes and characterizes the earth's terrestrial surface. It encompasses various types of ground, from natural landscapes such as forests, wetlands, and deserts to human-altered environments such as agricultural fields, urban areas, and industrial sites. LULC studies provide a snapshot of the earth's surface at a given time, offering valuable insights into the spatial distribution and interaction of various land use types and land cover classes. The dynamic nature of LULC, driven by both natural processes and human activities, necessitates continuous monitoring and analysis to capture temporal changes.

The importance of LULC studies extends to numerous fields. In environmental science, LULC data inform our understanding of biodiversity, ecosystem services, and the impacts of climate change. In urban planning and development, it helps to manage land resources, assess environmental impacts, and guide sustainable practices. LULC helps optimize land use for crop production in agriculture while minimizing environmental degradation. In addition, LULC data are integral to policy-making, supporting land conservation, urban growth, and climate change mitigation decisions.

Changes in Land Use Land Cover (LULC) offer insights into the dynamics of natural and human-altered landscapes, with applications spanning multiple disciplines. For example, LULC change data can elucidate the impacts of human activities on ecosystems, such as tracking deforestation rates to study biodiversity loss or carbon sequestration changes. LULC changes can also be used in climate change studies to influence local and global climates and inform climate modeling and future scenario prediction. LULC change data can inform the growth patterns of cities, aiding decisions about infrastructure development, zoning, and resource allocation. Monitoring LULC changes can optimize land use for crop production, identify shifts in agricultural practices, and assess environmental impacts. LULC change data also aids in disaster management, where changes in land cover, like deforestation, can increase vulnerability to natural disasters such as floods and landslides. Also, LULC changes inform policy decisions related to land conservation, urban growth, climate change mitigation, and sustainable development. 

Traditionally, remote sensing for LULC mapping involved manual or semi-automated image classification processes. While efficient in their time, these techniques were often time-consuming and meticulous tasks that required human intervention and expert knowledge. The manual interpretation was potentially prone to errors and discrepancies and lacked the scalability to cover large geographical areas.

AI, specifically machine learning (ML) algorithms, has significantly improved the ability to detect remote sensing in LULC studies. These algorithms learn from vast amounts of data, building models that can forecast future scenarios or identify patterns with exceptional accuracy. Recent progress in computer vision, a subfield of AI, has encouraged the development of sophisticated algorithms to teach computers how to see and analyze large amounts of image data, including satellite or aerial imagery used in remote sensing.

Among the classic Machine Learning methods for modeling LULC is K-nearest neighbor from \cite{thanh2017comparison} and \cite{qian2014comparing}, Random Forests by  \cite{thanh2017comparison}, \cite{ma2017review}, \cite{camargo2019comparative}, \cite{ghimire2012evaluation}, \cite{adam2014land}, \cite{shiraishi2014comparative}, \cite{raczko2017comparison}, and \cite{lee2015comparisons}, Support Vector Machines (SVM) have also been used in the works of \cite{thanh2017comparison}, \cite{qian2014comparing}, \cite{ma2017review}, \cite{camargo2019comparative}, \cite{shiraishi2014comparative}, \cite{raczko2017comparison}, \cite{srivastava2012selection}, \cite{pal2017detection}, and \cite{otukei2010land}. Other approaches such as Naive Bayes by \cite{camargo2019comparative}, Decision trees by \cite{thanh2017comparison}, \cite{qian2014comparing}, \cite{camargo2019comparative}, and \cite{otukei2010land}, Classic Neural Networks in the works of \cite{thanh2017comparison}, \cite{camargo2019comparative}, \cite{shiraishi2014comparative}, \cite{raczko2017comparison}, \cite{srivastava2012selection}, and \cite{pal2017detection}, Maximum likelihood classifiers by \cite{srivastava2012selection}, Bagging  in \cite{ghimire2012evaluation}, and Boosting in \cite{ghimire2012evaluation}.

More recently, Deep Learning-based models have taken the stage and dominated the remote sensing world with models based on Convolutional Neural Networks (CNNs) in the works of \cite{marcos2018land}, \cite{huang2018urban}, \cite{rezaee2018deep}, and \cite{shi2021actl}. Autoencoders have also been used to learn a representation of the data in \cite{zhou2015high}, \cite{azarang2019convolutional}, and \cite{hu2021hyperspectral}, Stacked Autoencoders by \cite{zhao2017spectral}, and \cite{sun2017encoding}, 3D Convolutional Autoencoders by \cite{mei2019unsupervised}, multitask deep learning in \cite{liu2020few}, Generative models by \cite{hong2020multimodal}, and Transformers by \cite{hong2021spectralformer}, \cite{xue2022deep}, \cite{qing2021improved}, \cite{jamali2022swin}, \cite{dong2021exploring}, and \cite{yao2023extended}.

This paper compares classification methods for land cover. We compare standard Convolutional Neural Network methods as well as state-of-the-art Transformer-based methods. We share all the code this project uses at \url{github.com/jrterven/eurosat_classification}.

\begin{figure*}
\includegraphics[width=\textwidth]{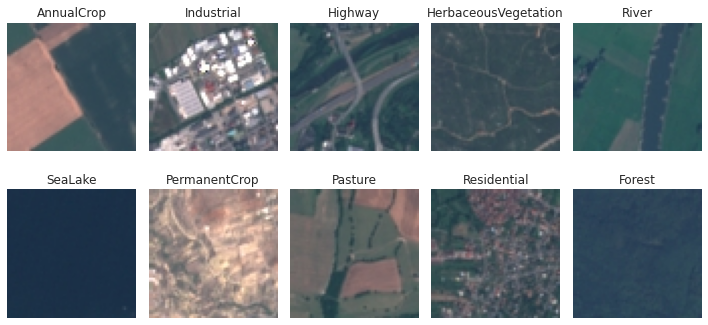}
\captionsetup{font=scriptsize}
\caption{Sample images from the EuroSAT dataset showing the ten categories.} 
\label{fig:eurosat_classes}
\end{figure*}

\section{The EuroSAT dataset}
Land Use Land Cover (LULC) categorizes and characterizes the earth's terrestrial surface. \emph{Land use} describes the human activities directly related to the land, indicating how people utilize land resources, such as residential, agricultural, or commercial. \emph{Land cover}, on the other hand, refers to the physical material at the surface of the earth, including grass, asphalt, trees, bare ground, water, etc. 

The EuroSAT dataset~\cite{EuroSAT} contains ten classes with 27000 labeled and geo-referenced images taken from Sentinel-2 satellite images. The images are $64 \times 64$ and cover cities in the European Urban Atlas. The covered cities are distributed over the 34 European countries: Austria, Belarus, Belgium, Bulgaria, Cyprus, Czech Republic (Czechia), Denmark, Estonia, Finland, France, Germany, Greece, Hungary, Iceland, Ireland, Italy, Latvia, Lithuania, Luxembourg, Macedonia, Malta, Republic of Moldova, Netherlands, Norway, Poland, Portugal, Romania, Slovakia, Slovenia, Spain, Sweden, Switzerland, Ukraine and United Kingdom. Figure \ref{fig:eurosat_classes} shows one random sample image for each of the ten classes. 

\begin{figure*}
\includegraphics[width=\textwidth]{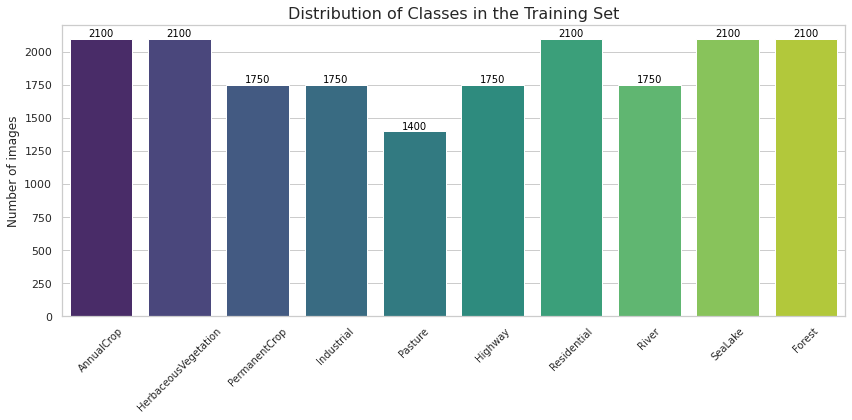}
\captionsetup{font=scriptsize}
\caption{Images distribution in the Training set.} 
\label{fig:train_dist}
\end{figure*}

The dataset provides split for training, validation, and testing with 18900, 5400, and 2700 images per split, respectively. Figure \ref{fig:train_dist} shows the distribution of images per class for the training set.

\section{Methods}

\subsection{Data Acquisition}
We obtained the dataset from the original source (\cite{phelber_eurosat_2023}), ensuring authenticity and data integrity. This dataset provides a collection of ten categories, mostly balanced (see Figure \ref{fig:train_dist}), necessary for training and evaluating our classification models.

\subsection{Data Preprocessing and Training}
We developed custom code for efficient data loading, preprocessing, and augmentation. The preprocessing steps included normalization and resizing of images. We trained ten models, seven convolutional architectures, and three transformer-based architectures. The convolutional models were AlexNet by \cite{krizhevsky2012imagenet}, ResNet50 by \cite{he2016deep}, ResNeXt by \cite{xie2017aggregated}, DenseNet by \cite{huang2017densely}, MobileNetV3 from \cite{howard2019searching}, EfficientNetV2 by \cite{tan2021efficientnetv2}, and ConvNeXt by \cite{liu2022convnet}. The transformed models were ViT by \cite{dosovitskiy2020image}, Swin Transformer by \cite{liu2021swin}, and MaxViT by \cite{tu2022maxvit}. 

We trained all models using an early stop of ten epochs and kept the model with the best validation accuracy. We used the categorical cross-entropy loss and the Adam optimizer by ~\cite{kingma2014adam} with a learning rate of $1\times10^{-4}$.

\subsubsection{Training from Scratch.}
Initially, we trained each of the ten models from scratch. This approach involved initializing the weights with the Kaiming Initialization proposed by ~\cite{he2015delving} and then retraining all layers using the RGB images of the EuroSAT dataset. The purpose of this step was to evaluate the learning capacity of each architecture without prior knowledge.

\subsubsection{Training with Pre-trained Weights.}
Subsequently, we employed transfer learning by training the same models using pre-trained weights. These weights were obtained from models pre-trained on ImageNet.

\subsection{Model Evaluation}
We evaluated each model on the test set, which was not used during the training phase. As evaluation metrics, we compute the Top-1 accuracy and the precision/recall curves. 

For our experiments, we used Pytorch and multiple computers: a Colab machine with NVIDIA V100 GPU, as well as two personal computers, one with TITAN X GPU and another equipped with RTX 4090 GPU.

\section{Results}
In this section, we present the results of the classification task using ten influential deep learning architectures trained from scratch and retrained with pre-trained parameters. Table \ref{table:accuracies} shows the Top-1 accuracy on the test set.  

\begin{table}[htbp]
\centering
\captionsetup{font=scriptsize}
\caption{Classification accuracy comparison on ten deep learning models. The \textit{Accuracy} column shows the results of the models trained from scratch, while the \textit{Accuracy pre-trained} column shows the results of retraining with pre-trained weights. The first seven are convolutional-based models, while the last three are Transformer-based models.}
\label{table:accuracies}
\vspace{8pt} 
\begin{tabular}{lcc}
\hline
\textbf{Model name}      & \textbf{Accuracy} & \textbf{Accuracy} \\ 
                         &                   & \textbf{pre-trained} \\ \hline
AlexNet                 & 0.837             & 0.916                          \\ 
ResNet                  & 0.835              & 0.947                          \\ 
ResNeXt                 & 0.843              & 0.981                          \\ 
DenseNet                & 0.925              & 0.949                          \\ 
MobileNetV3             & 0.877              & 0.958                          \\ 
EfficientNetV2          & 0.908              & 0.973                          \\ 
ConvNeXt                & 0.783             & 0.986                        \\ \hline
ViT32                  & 0.917              & 0.972                          \\ 
SwinB                  & 0.926              & 0.987                           \\ 
MaxViT                 & \textbf{0.973}     & \textbf{0.990}                  \\ \hline
\end{tabular}
\end{table}

\vspace{20pt}
As shown in the table, the model with the highest scores is also the latest architecture that we tried, achieving a state-of-the-art 99\% accuracy on this dataset. 

\begin{figure*}[ht!]
    \centering
    \begin{subfigure}{0.48\textwidth}
        \includegraphics[width=\linewidth]{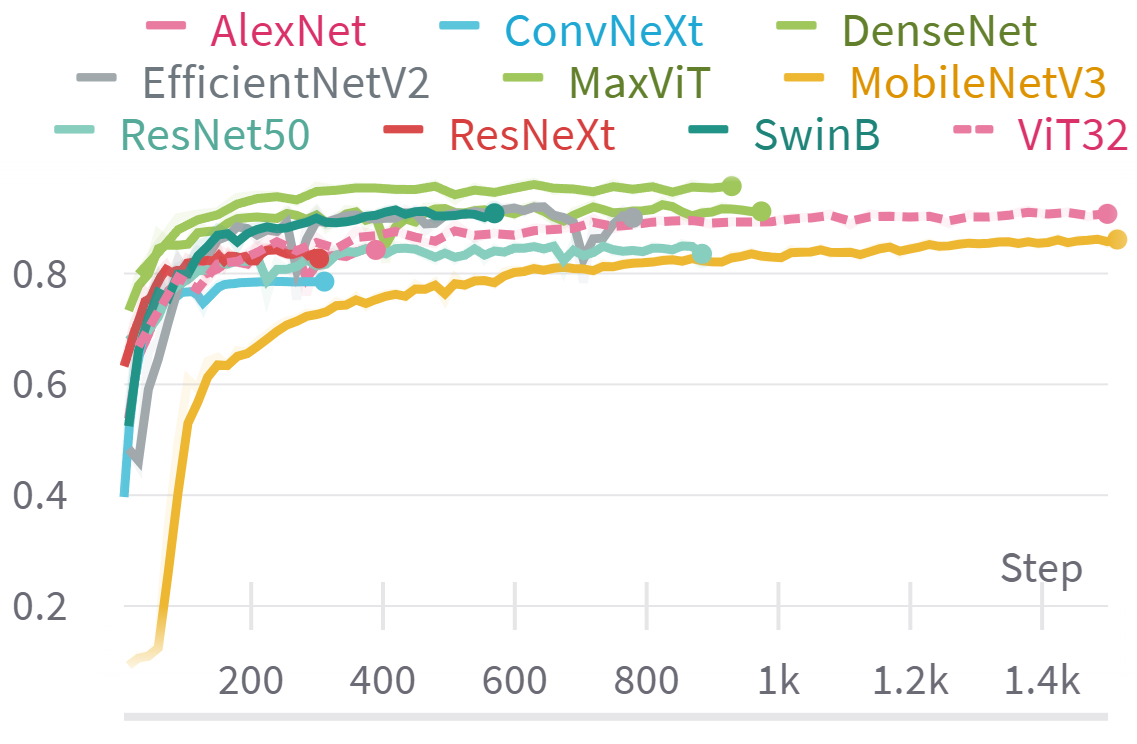}
        \captionsetup{font=small}
        \caption{Models trained from scratch}
        \label{fig:val_accuracy_from_scratch}
    \end{subfigure}
    \hfill
    \begin{subfigure}{0.48\textwidth}
        \includegraphics[width=\linewidth]{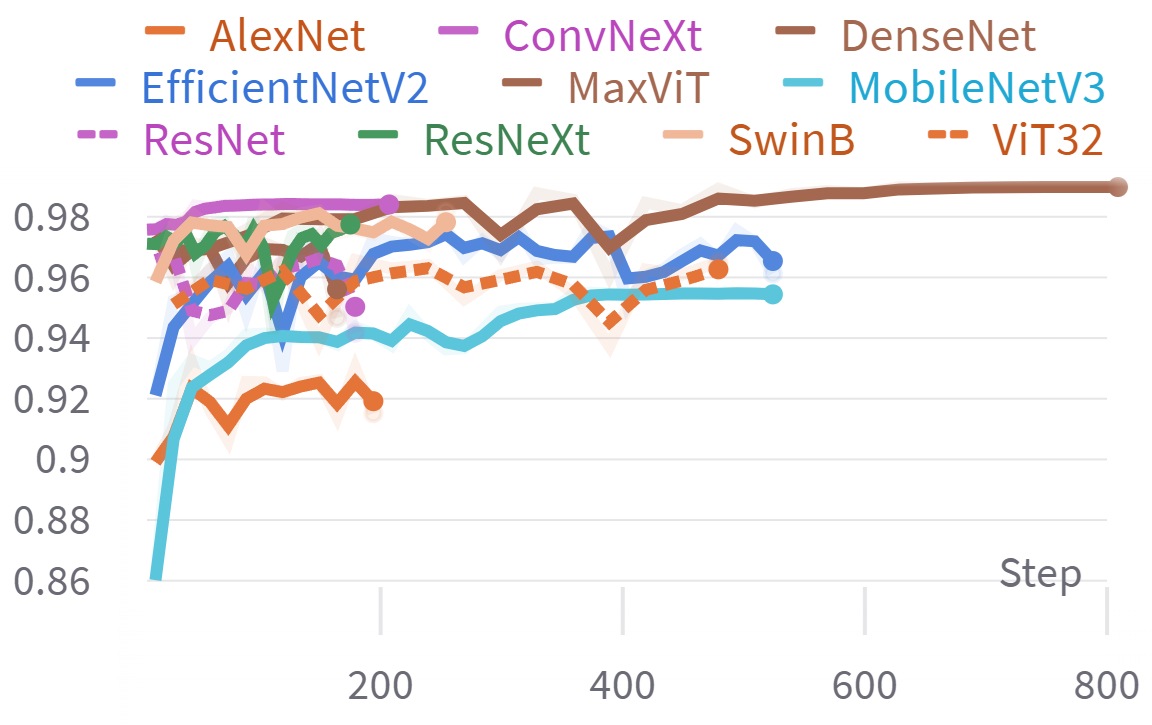}
        \captionsetup{font=small}
        \caption{Pretrained models}
        \label{fig:val_accuracy_pretrained}
    \end{subfigure}
    \captionsetup{font=scriptsize}
     \caption{Validation accuracy curves for the ten models. (a) Trained from scratch and (b) using pre-trained weights on ImageNet. Best viewed in color.}
\end{figure*}

Figures \ref{fig:val_accuracy_from_scratch} and \ref{fig:val_accuracy_pretrained} show the validation accuracy curves obtained during training. It should be noted that the models trained with random weights (Figure \ref{fig:val_accuracy_from_scratch}) took significantly longer to train and reached lower accuracy than the models trained with pre-trained weights (Figure \ref{fig:val_accuracy_pretrained}). We used the \cite{wandb} platform to log the experiments and generate these graphs. 

\begin{figure*}[ht!]
    \centering
    \begin{subfigure}{0.48\textwidth}
        \includegraphics[width=\linewidth]{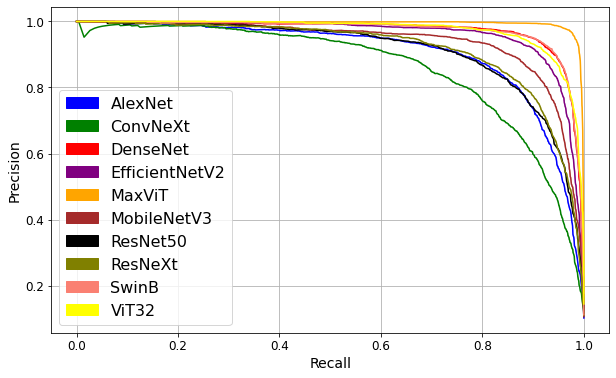}
        \captionsetup{font=small}
        \caption{Models trained from scratch}
        \label{fig:pr_from_scratch}
    \end{subfigure}
    \hfill
    \begin{subfigure}{0.48\textwidth}
        \includegraphics[width=\linewidth]{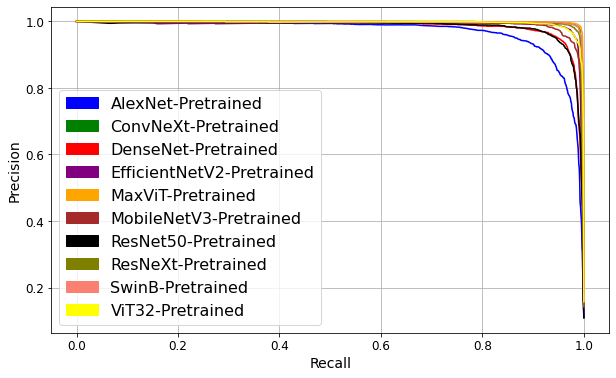}
        \captionsetup{font=small}
        \caption{Pretrained models}
        \label{fig:pr_pretrained}
    \end{subfigure}
    \captionsetup{font=scriptsize}
     \caption{Precision/Recall curves for the ten models. (a) Trained from scratch and (b) using pre-trained weights on ImageNet. Best viewed in color.}
\end{figure*}

Figures \ref{fig:pr_from_scratch} and \ref{fig:pr_pretrained} show the precision/recall curves for the ten models. The curves show that the models trained from scratch (Figure \ref{fig:pr_from_scratch}) present more variability in their results, with ConvNeXt on the lower end and MaxViT on the higher end. For the pre-trained models, the curves are closer to each other, with AlexNet as the lowest-performance model and MaxViT on the higher end. Curiously, ConvNeXt was the lowest-performance model trained from scratch but was on par with the best models when trained with pre-trained weights. This may indicate that the transfer learning used with the pre-trained weights helps alleviate the relatively small size of this dataset for such a model.

\section{Conclusion}
Land cover image classification is a crucial task in understanding environmental changes, urban planning, and disaster management. In this paper, we explored using state-of-the-art deep learning models for enhanced accuracy and efficiency in LC analysis. We compared convolutional neural networks (CNN) against transformer-based methods, illuminating their applications and advantages in LC studies. Our results showed that both CNN and transformer-based methods achieved high accuracy in LC classification, with transformer-based models outperforming CNNs in some cases. 

Using deep learning models in land cover classification offers a powerful avenue for research and applications. With the availability of large-scale remote sensing datasets and the rapid development of deep learning techniques, we can expect continued advancement in this field. As we move towards more automated and efficient methods for land cover classification, we can unlock the full potential of this data in addressing critical environmental and societal challenges.

\section{Acknowledgements}
We thank the Instituto Politecnico Nacional through the Research and Postgraduate Secretary (SIP) project number 20232290 and the Consejo Nacional de Humanidades Ciencias y Tecnologías (CONAHCYT) for its support through the Sistema Nacional de Investigadoras e Investigadores (SNII).

\bibliographystyle{ieeetr}
\bibliography{references}  

\end{document}